\documentclass[sigconf]{acmart}
\usepackage{enumitem}
\begin{document}

\title{Spatio-Temporal Trajectory Foundation Model: Recent Advances and Future Directions}

\author{Sean Bin Yang$^1$, Ying Sun$^2$, Yunyao Cheng$^1$, Yan Lin$^1$, Kristian Torp$^1$, Jilin Hu$^3$}
\affiliation{%
  \institution{$^1$Aalborg University, $^2$ Chongqing University of Posts and Telecomunications, $^3$East China Normal University}
  \country{}
}
\email{{seany, yunyaoc, lyan, torp}@cs.aau.dk, sunying@cqupt.edu.cn, jlhu@dase.ecnu.edu.cn}

%

%
%
%
%



\begin{abstract}
Foundation models (FMs) have emerged as a powerful paradigm, enabling a diverse range of data analytics and knowledge discovery tasks across scientific fields. Inspired by the success of FMs—particularly large language models—researchers have recently begun to explore spatio-temporal foundation models (STFMs) to improve adaptability and generalization across a wide spectrum of spatio-temporal (ST) tasks. Despite rapid progress, a systematic investigation of trajectory foundation models (TFMs), a crucial subclass of STFMs, is largely lacking. This tutorial addresses this gap by offering a comprehensive overview of recent advances in TFMs, including a taxonomy of existing methodologies and a critical analysis of their strengths and limitations. In addition, the tutorial highlights open challenges and outlines promising research directions to advance spatio-temporal general intelligence through the development of robust, responsible, and transferable TFMs.

\end{abstract}

\maketitle

\section{Introduction}
Self-supervised learning (SSL), such as contrastive learning \cite{DBLP:conf/ijcai/YangGHT021,DBLP:conf/icde/YangGHYTJ22,DBLP:conf/kdd/0001CGGHY025,DBLP:conf/www/Wei0G0YH25}, and generative learning \cite{DBLP:conf/kdd/YangHGYJ23,DBLP:journals/pvldb/ZhouSCJK24,DBLP:conf/cikm/ChenLCBLLCE21,DBLP:journals/tkde/LinZLLWLLJGLW25,DBLP:conf/aaai/LinW0L21}, has exhibited strong theoretical foundations and remarkable empirical performance across a wide range of domains, including natural language processing \cite{DBLP:conf/naacl/DevlinCLT19}, computer vision \cite{DBLP:conf/eccv/TianKI20}, and graph data \cite{DBLP:conf/nips/YouCSCWS20}. These SSL approaches enable models to learn high-quality representations from large volumes of unlabeled data, thereby reducing reliance on costly manual annotations and improving generalization in downstream tasks. 

\begin{figure}[!t]
	\centering
	\includegraphics[scale=0.47]{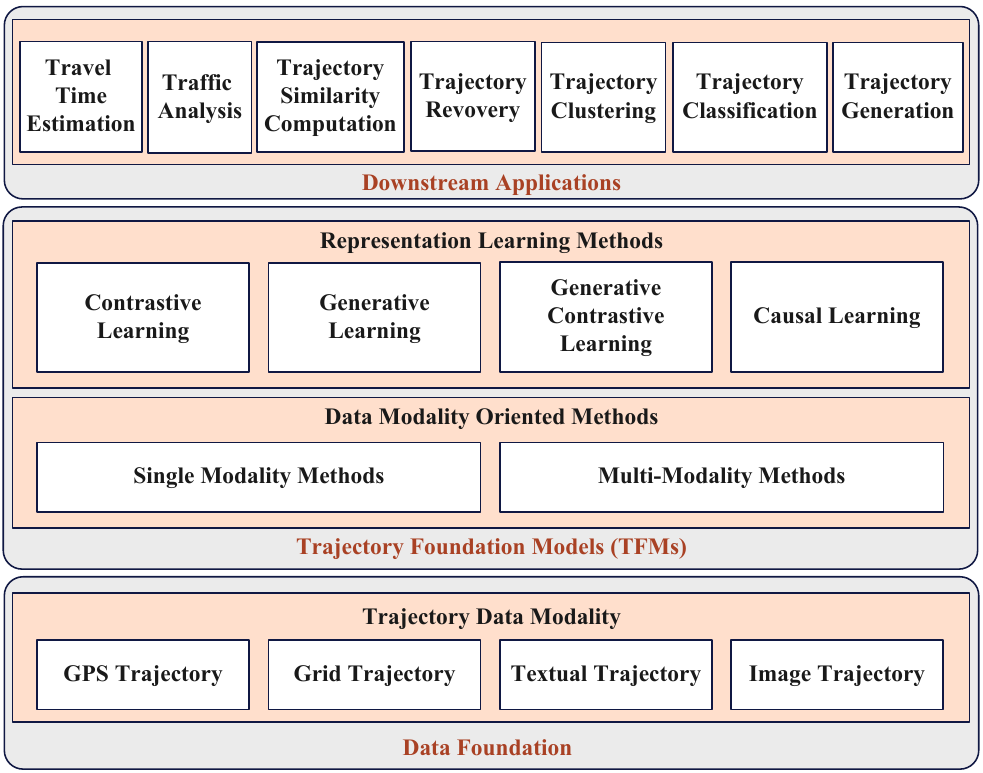}
	\caption{The framework of TFMs for ST Trajectory data.}
	\label{fig:Example}
	\vspace{-10pt}
\end{figure}
As shown in Figure \ref{fig:Example}, trajectory foundation models (TFMs)—also referred to in earlier literature as trajectory representation learning—serve as a fundamental enabler for a variety of intelligent transportation applications, including travel time estimation \cite{DBLP:conf/ijcai/YangGHT021,DBLP:conf/icde/YangGHYTJ22,DBLP:conf/kdd/0001CGGHY025,DBLP:conf/www/Wei0G0YH25,DBLP:conf/kdd/YangHGYJ23,DBLP:journals/pvldb/ZhouSCJK24,DBLP:conf/cikm/ChenLCBLLCE21,DBLP:conf/icde/JiangPRJLW23,DBLP:journals/pacmmod/LinWHGYLJ23,DBLP:conf/gis/WauryJT19}, 
 traffic analysis \cite{DBLP:journals/pvldb/FangPCDG21,DBLP:conf/www/MaTCZXZCZG24,DBLP:conf/aaai/HanWWYLLW25,DBLP:conf/aaai/00220ZDL25,DBLP:conf/ijcai/FangWPCG22,DBLP:journals/tkde/YangGY22,DBLP:conf/icde/Yang020,DBLP:conf/icde/JensenYGHT24}, trajectory similarity computation \cite{DBLP:journals/tkde/HuCFFLG24,DBLP:conf/kdd/ZhouS00J25,GTR11,DBLP:conf/www/MaTCZXZCZG24,DBLP:conf/icde/Chang0LT23,DBLP:conf/cikm/DengZFSL022,DBLP:conf/kdd/Fang0ZHCGJ22,TrajCogn11,DBLP:journals/tkde/LinWGHJL24,HaoICDE25,DBLP:conf/mdm/00070L24}, trajectory recovery \cite{DBLP:journals/tkde/LinHGYJLW25,GTR11}, trajectory clustering \cite{DBLP:journals/tits/WangHWLJJWW24,DBLP:conf/icde/FangDCHGC21,DBLP:conf/icde/0001WCGHB24,DBLP:journals/tbd/SiYXLTZ25}, trajectory classification \cite{DBLP:conf/icde/YangGHYTJ22,DBLP:conf/cikm/LiangOWLCZZZ22,DBLP:journals/pvldb/ZhouSCJK24,DBLP:conf/kdd/ZhouS00J25,TrajRL,DBLP:conf/ijcai/LuoZ0WZRL24}, and trajectory generation \cite{GTR11,DBLP:conf/kdd/ZhuY0LY00WL24,DBLP:conf/nips/ZhuYZZY23}. In particular, a trajectory can be formally defined as $T=\langle(x_1,y_1,t_1),(x_2,y_2,t_2),\cdots, (x_n,y_n,t_n)\rangle$, where each element represents a spatio-temporal point consisting of spatial coordinates $(x_i, y_i)$ and a corresponding timestamp $t_i$. Such trajectories inherently capture both spatial movement patterns and temporal dynamics, making them a rich yet challenging modality for effective representation learning.

\noindent
\textbf{Tutorial Overview. } With the rapid proliferation of spatio-temporal data, trajectory foundation models have emerged as a critical research direction, attracting increasing attention from both academia and industry. This tutorial provides a systematic and comprehensive overview of this evolving field. We begin by formalizing the core concepts of trajectory data mining and trajectory foundation models, establishing a unified basis for subsequent discussions. The objectives of this tutorial are threefold: (i) to elucidate the motivations for developing trajectory foundation models by analyzing the diverse modalities of trajectory data and their unique characteristics; (ii) to review and critically examine recently proposed approaches from a modality-oriented perspective, with particular emphasis on the challenges, limitations, and open problems associated with each modality; and (iii) to investigate the emerging paradigm of trajectory foundation model construction, presenting a structured taxonomy of learning methodologies and offering analytical insights into their theoretical foundations and practical implications. The tutorial is organized as follows: (1) \textbf{Introduction, Motivation, and Foundation (10 mins)}. (2) \textbf{Data-Modality-Oriented Methods (20 mins)}. (3) \textbf{Foundation Model Learning Methods (50 mins)}. (4) \textbf{Conclusion and Future Directions (10 mins)}.

This tutorial features: (i) an extensive review of the literature from the past five years, encompassing both pioneering contributions and the latest state-of-the-art approaches, as developed across multiple research communities ranging from data management to machine learning; (ii) insights into recent advancements across diverse solution strategies, thereby enriching the research landscape with novel perspectives and methodological innovations. This tutorial will equip the audience with a fundamental comprehension of TFMs and inspire them to work in this dynamic and evolving area.

\noindent
\textbf{Relation to Previous Tutorials. }Recent tutorials related to spatio-temporal foundation models have primarily focused (i) on broader aspects of spatio-temporal data—such as time series, video, and trajectories—while overlooking several state-of-the-art methods specific to trajectory data \cite{DBLP:journals/corr/abs-2503-13502,DBLP:journals/corr/abs-2503-08473}; (ii) on specific types of spatial-temporal data (such as time series \cite{DBLP:conf/kdd/LiangWNJ0SPW24,DBLP:journals/corr/abs-2507-10620}).

In contrast, this tutorial offers a substantial expansion over prior efforts, providing a detailed exposition of recently proposed methodologies complemented by interactive exploration for the audience. Beyond introducing techniques for trajectory foundation models, the tutorial adopts a comprehensive perspective on the field, emphasizing the diversity of trajectory data modalities and the advancement of learning methods. A section introduces a novel taxonomy of TFMs, thereby opening up promising research avenues. To the best of our knowledge, this is the first tutorial that (i) presents an extensive coverage of the trajectory foundation model landscape; (ii) integrates the most recent developments in foundation models for spatio-temporal trajectory data; and (iii) outlines future directions informed by large-scale trajectory datasets and principles of responsible artificial intelligence (AI), with particular attention to sustainability-oriented AI.

\noindent
\textbf{Audience and Expected Background. }This tutorial is designed for researchers and data analysts, with a particular focus on recent advances in TFMs. It aims to foster new collaborations between members of the data management community and data science practitioners from diverse application domains, while simultaneously broadening interest in TFMs. The tutorial provides the necessary background to ensure that participants can follow the entire presentation, alongside in-depth technical discussions of state-of-the-art solutions. In addition, we critically examine the limitations of existing approaches and highlight key open research problems. The material is structured to be accessible to newcomers, yet sufficiently detailed to offer valuable insights for experts in the field.
\vspace{-5pt}
\section{Tutorial Outline}
In this 1.5-hour lecture-style tutorial, we systematically examine the problem of foundation models in the trajectory domain, beginning with the fundamental definitions of trajectory data modalities and trajectory foundation models, and progressing to the open challenges and emerging opportunities introduced by the development of responsible TFMs. 


\subsection{Introduction, Motivations and Foundations}


We begin by discussing representative scientific and industrial applications that critically depend on the effectiveness of trajectory foundation models. 


\textbf{Type of Trajectory Data Modality. }We then introduce the different types of trajectory data modalities. Specifically, we categorize trajectory data into four primary modalities, each defined by its unique characteristics: (i) \textbf{GPS Trajectory: }The GPS trajectory $T$ represents a sequence of GPS points collected at a fixed sampling time interval. Each point is defined as $T_i=(x_i,y_i,t_i)$, where $x_i$ and $y_i$ are the longitude and latitude, respectively, and $t_i$ is the corresponding timestamp. (ii) \textbf{Grid Trajectory: }The grid trajectory is a sequence of timestamped grid cells obtained by transforming GPS points in $T$ into different grids, where each cell is formulated as $\tau_i^{g}=(g_i,t_i^{g})$, where $g_i$ is the ID of a grid and $t_i^{g}$ is the time; (iii) \textbf{Textual Trajectory: }The textual trajectory represents a sequence of timestamped textual descriptions of each spatial location $(x_i,y_i)$, such as place names, points of interest, or functional categories (e.g., 'restaurant', 'residential area'); (iv) \textbf{Image Trajectory: }A trajectory represented as a sequence of timestamped satellite images, where each image corresponds to a GPS location in $T$. Formally, each element augments the spatio-temporal point $(x_i,y_i,t_i)$ with an associated image patch $I_i$, thereby providing a rich visual context for the surrounding environment along the trajectory.

\textbf{Type of Downstream Applications. }We then introduce the different types of downstream applications (see Figure \ref{fig:Example}). \textbf{Travel time estimation} refers to the task of predicting the time required to traverse a given trajectory $T_i$. \textbf{Traffic Analysis} aims to predict and characterize traffic flow within a specific region, typically by modeling the spatio-temporal dynamics of vehicle movements and road network conditions. \textbf{Trajectory Similarity Computation} refers to the task of retrieving the top-$k$ most similar trajectories from a large-scale dataset, typically based on spatio-temporal proximity, semantic attributes, or learned representation spaces. \textbf{Trajectory Recovery} aims to infer and reconstruct missing spatial locations within a trajectory, thereby restoring incomplete movement records and enhancing the quality of spatio-temporal data for downstream analysis. \textbf{Trajectory Clustering} refers to the task of grouping trajectories into clusters based on their spatial, temporal, or semantic similarity, with the goal of uncovering common movement patterns, identifying representative behaviors, and supporting applications such as traffic management and urban planning. \textbf{Trajectory Generation} aims to synthesize realistic trajectories that capture the spatial, temporal, and semantic patterns of real-world movement, often leveraging TFMs and generative models to support applications such as data augmentation,  and simulation.

\vspace{-5pt}
\subsection{Data-Modality-Oriented Methods}
We dive into the diverse TFMs proposed in the literature, systematically organized by their underlying data modalities. The discussion encompasses both single-modality approaches and multi-modality frameworks, with an emphasis on their methodological characteristics, comparative advantages, and practical implications for trajectory-related applications.

\textbf{Single Modality Methods. }Single-modality methods \cite{DBLP:conf/ijcai/YangGHT021,DBLP:journals/pvldb/ZhouSCJK24,DBLP:conf/icde/YangGHYTJ22,DBLP:conf/icde/JiangPRJLW23,DBLP:journals/tkde/LinZLLWLLJGLW25,TrajRL,DBLP:conf/ijcai/LuoZ0WZRL24,DBLP:conf/aaai/HanWWYLLW25} focus on leveraging GPS trajectories—either in the form of raw GPS point sequences or map-matched sequences of road segments—to learn informative and transferable trajectory representations. For example, PIM \cite{DBLP:conf/ijcai/YangGHT021} is among the earliest works that learn generic path representations from map-matched spatial trajectories, demonstrating strong generalization performance on tasks such as travel time estimation and path ranking. In contrast, RED \cite{DBLP:journals/pvldb/ZhouSCJK24} incorporates both spatial and temporal features to learn robust trajectory representations, and it evaluates the resulting embeddings across multiple downstream applications, including trajectory classification and trajectory similarity computation. In summary, by concentrating solely on GPS-derived data, single-modality methods aim to extract intrinsic movement patterns that can generalize across tasks, thereby providing an efficient foundation for a variety of downstream applications.
Despite their utility, single-modality methods face key limitations. Raw GPS trajectories are often noisy, sparse, and temporally irregular, while map-matched sequences introduce errors from imperfect matching and omit contextual factors such as traffic or environment. Relying solely on a single modality also restricts the ability to capture complex mobility behaviors, reducing robustness and generalizability in real-world applications.

\textbf{Multi-Modality Methods. }Multi-modality methods \cite{DBLP:conf/kdd/0001CGGHY025,DBLP:conf/www/Wei0G0YH25,GTR11,DBLP:conf/www/MaTCZXZCZG24,DBLP:conf/kdd/ZhouS00J25,DBLP:conf/aaai/00220ZDL25} seek to leverage two or more complementary data modalities—such as GPS trajectories combined with grid-based trajectories, textual trajectories, or image trajectories—to enhance trajectory foundation models. By integrating heterogeneous sources of information, these approaches enable the capture of richer spatial, temporal, and semantic characteristics, thereby overcoming the limitations of single-modality methods and improving generalization across diverse applications. For example, inspired by advances in computer vision, MM-Path \cite{DBLP:conf/kdd/0001CGGHY025} proposes a spatial trajectory representation learning framework that integrates GPS trajectories with image trajectories. By incorporating visual context from satellite images into the representation learning process, MM-Path achieves superior performance compared to earlier single-modality methods such as PIM. Meanwhile, Path-LLM \cite{DBLP:conf/www/Wei0G0YH25} explores the integration of GPS trajectories with textual trajectories to learn generic spatial trajectory representations. Leveraging the capabilities of large language models, Path-LLM aligns spatial movements with semantic descriptions, thereby enabling more expressive and transferable trajectory representations. However, current models primarily focus on two modalities, which constrains their generalization ability and limits their applicability across more complex real-world scenarios.

\vspace{-5pt}
\subsection{Foundation Model Learning Methods}
After presenting the various data-modality-oriented methods, we turn to foundation model learning approaches, examining how different learning paradigms contribute to the development of trajectory foundation models.

\textbf{Contrative Learning. }Contrastive learning-based TFMs \cite{DBLP:conf/aaai/00220ZDL25,TrajRL,DBLP:conf/www/Wei0G0YH25,DBLP:conf/kdd/0001CGGHY025,DBLP:conf/icde/YangGHYTJ22,DBLP:conf/ijcai/YangGHT021} aim to learn generic trajectory representations by constructing multiple views of the same trajectory based on contrastive objectives. The contrastive objective encourages the representations of positive trajectory views to be pulled closer in the embedding space, while simultaneously pushing them apart from negative trajectory views, thereby enhancing discriminability and generalization. For example, TrajRL \cite{TrajRL} employs trajectory augmentation to generate two views for self-supervised pre-training with multiple supervisory signals, and applies a contrastive objective to learn generic trajectory representations that incorporate multi-faceted temporal features. In contrast, MM-Path    
\cite{DBLP:conf/kdd/0001CGGHY025} introduces image trajectories to construct positive samples, thereby avoiding the need for pre-defined view generation. A contrastive objective is then applied between GPS and image trajectories to learn more generalized trajectory representations.
However, contrastive learning–based TFMs primarily capture global trajectory information while often overlooking fine-grained local trajectory patterns.

\textbf{Generative Learning. }Generative learning-based TFMs \cite{DBLP:conf/cikm/ChenLCBLLCE21,DBLP:journals/pvldb/ZhouSCJK24,GTR11} aim to pre-train TFMs by reconstructing or generating parts of the input trajectory from corrupted or masked versions. For TFMs, we typically involve tasks such as masked trajectory recovery, future trajectory prediction, or cross-modal generation, enabling models to learn richer spatio-temporal representations than purely contrastive approaches. For example, RED \cite{DBLP:journals/pvldb/ZhouSCJK24} employs a Transformer-based Masked Autoencoder (MAE) with road-aware masking and spatial-temporal-user joint embeddings to fully exploit trajectory information, yielding significant accuracy gains. In contrast, Toast \cite{DBLP:conf/cikm/ChenLCBLLCE21} adopts a generative self-supervised strategy by masking consecutive sequences of road segments and training the model to reconstruct the masked segments, thereby enabling the construction of transferable TFMs. However, these methods primarily capture local trajectory information, while often overlooking global trajectory dependencies and long-range mobility patterns.

\textbf{Generative Contrastive Learning. }As discussed earlier \cite{DBLP:conf/aaai/00220ZDL25,TrajRL,DBLP:conf/www/Wei0G0YH25,DBLP:conf/kdd/0001CGGHY025,DBLP:conf/icde/YangGHYTJ22,DBLP:conf/ijcai/YangGHT021,DBLP:conf/cikm/ChenLCBLLCE21,DBLP:journals/pvldb/ZhouSCJK24,GTR11}, contrastive learning primarily focuses on capturing the global contextual information of trajectories, whereas generative learning emphasizes modeling the local structural details. Building on this complementarity, several recent studies \cite{DBLP:conf/aaai/00220ZDL25,DBLP:conf/www/MaTCZXZCZG24,DBLP:conf/kdd/ZhouS00J25,DBLP:conf/kdd/YangHGYJ23,DBLP:conf/icde/JiangPRJLW23} have proposed to integrate generative and contrastive objectives. Such hybrid approaches effectively exploit both global and local information, thereby enhancing the generalization ability of TFMs and enabling the learning of more robust trajectory representations. For example, LightPath \cite{DBLP:conf/kdd/YangHGYJ23} introduces a lightweight and scalable framework for spatial trajectory representation. It employs a random masking strategy to adjust the input trajectory length and effectively capture local structural information, while incorporating relation reasoning to model the global contextual dependencies of spatial trajectories, thereby enabling the learning of more generic and transferable trajectory representations. START \cite{DBLP:conf/icde/JiangPRJLW23} integrates span-masked trajectory recovery with trajectory contrastive learning to capture rich spatio-temporal information. To enhance the effectiveness of contrastive learning, it further introduces four trajectory data augmentation strategies, thereby enabling more robust and discriminative trajectory representations. However, these hybrid approaches also face limitations. Balancing generative and contrastive objectives remains challenging, and most methods are confined to a small set of modalities, limiting their ability to capture the heterogeneity of real-world trajectories. In addition, scalability to large, noisy datasets is still underexplored, highlighting the need for more adaptive and efficient hybrid frameworks for TFMs.

\textbf{Causal Learning.} TFMsaim to characterize human movement behavior and represent a pivotal step toward understanding mobility patterns. However, existing methods frequently overlook the confounding influence of geospatial context, which can induce spurious correlations and compromise generalization. To mitigate this issue, recent advances in causal representation learning \cite{DBLP:conf/icde/Li0GCJZZFB24,DBLP:conf/ijcai/LuoZ0WZRL24} integrate deep learning with causal inference, thereby enabling the derivation of more robust and interpretable trajectory representations. For instance, TrajCL \cite{DBLP:conf/ijcai/LuoZ0WZRL24} introduces a causal-learning-based trajectory representation framework that formulates a Structural Causal Model and applies backdoor adjustment to eliminate spurious correlations arising from geospatial context, ultimately enhancing generalization, interpretability, and performance in trajectory classification tasks.

\subsection{Conclusion}



We conclude the tutorial by synthesizing the key insights derived from recent advances in foundation models and pre-training methodologies. Finally, we highlight promising future research directions and pressing challenges that arise with the advent and development of TFMs. Particular emphasis is placed on the role of large-scale, multi-modal trajectory datasets in driving progress, as well as the necessity of developing responsible and trustworthy TFMs aligned with societal values. Beyond technical advances, we underscore the importance of responsible AI principles, including fairness, transparency, and accountability, to ensure that TFMs are deployed ethically and equitably in real-world mobility systems. In addition, we discuss how trajectory foundation models can contribute to broader sustainability goals, such as enabling greener urban mobility, supporting climate resilience, and informing environmentally conscious decision-making. Looking ahead, we explore emerging opportunities at the intersection of TFMs and quantum AI, where quantum-inspired learning paradigms and scalable quantum computing techniques may unlock new capabilities for spatio-temporal modeling, offering both efficiency gains and fundamentally novel perspectives for building the next generation of responsible and sustainable trajectory foundation models.
\subsection{Future Directions}
We present a list of promising future research directions of the topic of TRL. 
Here, we show three directions of lack of space.

\textbf{Quantum Trajectory Data Representation. }
The explosive growth of trajectory data in spatio-temporal applications has made large-scale model training increasingly resource-intensive and time-consuming. However, the continuous scaling of computational power is constrained by fabrication bottlenecks and heat dissipation challenges, creating significant obstacles to further progress. Quantum computing, with its intrinsic superposition and parallelism, offers a promising alternative by enabling compact yet expressive trajectory representations that are efficiently manipulable. When combined with advances in quantum artificial intelligence, such representations extend beyond efficient encoding, supporting quantum-enhanced learning paradigms capable of accelerating tasks such as classification, clustering, and prediction. This synergy allows trajectory models not only to achieve superior scalability and energy efficiency, but also to explore novel forms of representation learning that are unattainable in classical settings.

\textbf{Responsible Foundation Models for Trajectories. }
Foundation models have greatly advanced trajectory representation by capturing generalizable spatio-temporal patterns across diverse contexts. However, the increasing scale of these models raises critical concerns about fairness, transparency, and environmental impact. A responsible approach emphasizes unbiased learning, interpretability, and robustness, ensuring that trajectory representations remain reliable and trustworthy across heterogeneous datasets. At the same time, the demand for sustainability calls for reducing energy consumption and carbon emissions during model training and deployment. Techniques such as efficient architectures, knowledge distillation, and green optimization provide promising pathways toward eco-friendly foundation models. 

\textbf{Continued Pre-training of TFMs. }
Pre-training is a powerful paradigm for TFMs, enabling models to learn rich spatio-temporal patterns from large-scale unlabeled data. Yet, most current approaches are static, relying on a one-off pre-training process that struggles to adapt when new data or domains emerge. Continued pre-training addresses this limitation by incrementally incorporating new data while retaining previously acquired knowledge. This incremental approach mitigates catastrophic forgetting and allows models to remain effective in dynamic mobility environments with evolving patterns and heterogeneous sources. By incorporating techniques such as continual learning, adaptive fine-tuning, and parameter-efficient updates, continued pre-training provides a scalable path toward lifelong trajectory intelligence. 

\section{Presenters}

\noindent
\textbf{Sean Bin Yang} is a tenure-track Assistant Professor at Aalborg University, Denmark. His research interests mainly focus on the areas of spatio-temporal data mining, representation learning, and artificial intelligence over large scale data. Most of his research works were published in top-tier conferences and journals (e.g., KDD, ICDE, and TKDE).

\noindent
\textbf{Ying Sun} is a tenure-track Assistant Professor at Chongqing University of Posts and Telecommunications, working on spatiotemporal data intelligence, construction waste recognication, generative AI, and AI for environments.

\noindent
\textbf{Yunyao Cheng} is a postdoctoral researcher at Aalborg University. His reserach interests mainly lie in time series data mining. His research works were often publised in top-tier database and AI conferences (e.g., PVLDB, and ICLR).

\noindent
\textbf{Yan Lin} is an Assistant Professor at Aalborg University. His reserach interests focus on spatio-temporal data mining. His research works were often published in top-tier database and AI conferences and Journals (e.g., SIGMOD, AAAI, and TKDE).

\noindent
\textbf{Kristian Torp} is a professor at Aalborg University. His research includes\textbf{} aspects of managing and querying large spatio-temporal datasets. He has contributed ideas and software to multiple products used in the transportation domain.

\noindent
\textbf{Jilin Hu} is a Professor at East China Normal University. His research focuses on time series analysis and spatio-temporal data management. He has authored numerous publications in top-tier venues, including leading database and data mining conferences (e.g., SIGMOD, and PVLDB) and premier machine learning conferences (e.g., ICML, and ICLR).

\section{Acknowledgments}
The paper has received fundings from the European Union’s funded Project MobiSpaces (101070279), National Secience and Foundation of China (62472174, 62402082), Chongqing Municipal Education Commission's Key Project in Humanities and Social Sciences (23SKJD07), and Scientific and Technological Research Program of Chongqing Municipal Education Commission (KJQN202400637).


\balance
\bibliographystyle{ACM-Reference-Format}
\bibliography{sample-base}

\end{document}